\documentclass[runningheads]{llncs}
\usepackage[T1]{fontenc}
\usepackage{graphicx}
\usepackage{multirow}
\usepackage{makecell}
\usepackage[misc]{ifsym}
\begin{document}
\title{View Distribution Alignment with Progressive Adversarial Learning for UAV Visual Geo-Localization
	\thanks{This work was supported in part by the National Natural Science Foundation of China under Grant No. 62171295, and in part by the Liaoning Provincial Natural Science Foundation under Grant No.2021-MS-266, and in part by the Applied Basic Research Project of Liaoning Province under Grant 2023JH2/101300204, and in part by the Shenyang Science and Technology Innovation Program for Young and Middle-aged Scientists under Grant No.RC210427.}}
\titlerunning{View Distribution Alignment for UAV Visual Geo-Localization}

\author{Cuiwei Liu \and Jiahao Liu\and Huaijun Qiu \textsuperscript{\Letter} \and Zhaokui Li \and Xiangbin Shi}

\authorrunning{C. Liu et al.}


\institute{School of Computer Science, Shenyang Aerospace University, Shenyang, China \\
\email{Corresponding author: Huaijun Qiu}\\
\email{Email: liucuiwei@sau.edu.cn; mraliens@163.com; 20220071@email.sau.edu.cn; lzk@sau.edu.cn; sxb@sau.edu.cn}
}

\maketitle              
\begin{abstract}
Unmanned Aerial Vehicle (UAV) visual geo-localization aims to match images of the same geographic target captured from different views, i.e., the UAV view and the satellite view.
It is very challenging due to the large appearance differences in UAV-satellite image pairs.
Previous works map images captured by UAVs and satellites to a shared feature space and employ a classification framework to learn location-dependent features while neglecting the overall distribution shift between the UAV view and the satellite view.
In this paper, we address these limitations by introducing distribution alignment of the two views to shorten their distance in a common space.
Specifically, we propose an end-to-end network, called PVDA (Progressive View Distribution Alignment).
During training, feature encoder, location classifier, and view discriminator are jointly optimized by a novel progressive adversarial learning strategy.
Competition between feature encoder and view discriminator prompts both of them to be stronger.
It turns out that the adversarial learning is progressively emphasized until UAV-view images are indistinguishable from satellite-view images.
As a result, the proposed PVDA becomes powerful in learning location-dependent yet view-invariant features with good scalability towards unseen images of new locations.
Compared to the state-of-the-art methods, the proposed PVDA requires less inference time but has achieved superior performance on the University-1652 dataset.

\end{abstract}
\section{Introduction}

Cross-view geo-localization task is to acquire the real-world geographic position of a given image by retrieving the most relevant images in a geo-tagged reference database captured from another view, e.g., the satellite view.
Such technologies have attracted great attention since they are particularly useful in practical applications, such as autonomous driving~\cite{rsObject2022}, augmented reality~\cite{AR2019}, and mobile robots~\cite{2012SatelliteRobot}.
Cross-view geo-localization was first presented to address ground-to-aerial geo-localization task~\cite{lin2015learning,zhai2017CVUSA,tian2017cross,liu2019CVACT}, which matches a ground-view query image against aerial-view reference images with geo-tags.
Recently, some works~\cite{zheng2020university} consider the UAV visual geo-localization problem as bidirectional cross-view matching between UAV-view images and satellite-view images.
Specifically, UAV-to-satellite image matching achieves UAV-view target localization in GPS-denied situations by comparing a UAV-view query with a collection of geo-tagged satellite-view images.
Conversely, satellite-to-UAV image matching navigates the UAV back to a target place in the satellite-view image by searching for the most similar UAV-view images in the flight historic record.

UAV visual geo-localization is essentially a cross-view scene image retrieval task based on the characteristics of the geographic target with its surroundings, such as neighbor houses, roads, and trees.
The variations in viewpoint, height, and seasons lead to large appearance differences in UAV-satellite image pairs and pose great challenges for accurate image matching.
The mainstream methods utilize a classifier of different geographic locations as a proxy to train a common feature space for images captured by UAVs and satellites, then the learned space is used at inference to extract descriptors for image retrieval.
Early studies~\cite{zheng2020university,ding2020practical} extract global features from the whole image, while more appealing works~\cite{wang2021each,lin2022joint,zhuang2021faster,dai2022transformer} take spatial or semantic contextual patterns into consideration and learn local features for part matching.
However, the above works~\cite{zheng2020university,ding2020practical,wang2021each,zhuang2021faster,dai2022transformer} neglect the significant appearance gap between UAV-view images and satellite-view images, leading to inferior image retrieval performance when applying the learned feature space to unseen images of new locations.
Some recent cross-view geo-localization methods for ground-to-aerial geo-localization~\cite{2019Bridging,2019ShiNIPS,2022SoftExemplar,2022SSA-Net} and UAV visual geo-localization~\cite{tian2021uav} employ cross-view image synthesis techniques as data augmentation to explicitly transform images from one view to another view before feature extraction.
Although such data augmentation techniques narrows down the gap between query and reference images, they require a two-step procedure for image retrieval and put an extra burden on computation time as well as resources for practical applications.

Unlike previous works, we handle the domain shift problem between UAV-view images and satellite-view images by performing distribution alignment of these two views in a common feature space to narrow the gap between them.
We propose an end-to-end network, called PVDA (Progressive View Distribution Alignment), which pulls UAV-view images and satellite-view images of the same location together considering both view distribution alignment and location classification.
We introduce a view discriminator to determine whether an image was captured by a UAV or a satellite.
A novel progressive adversarial learning strategy is designed to jointly optimize the feature encoder, the location classifier and the view discriminator in a unified framework.
The feature encoder is optimized under the guidance of the location classifier while trying to fool the view discriminator into regarding UAV-view images as satellite-view images and vice versa.
It is increasingly hard to deceive the view discriminator over time since competition between the feature encoder and the view discriminator prompts both of them to be stronger.
To solve this problem, the proposed learning strategy progressively emphasizes the task of confusing the view discriminator and simulates a warm restart of the learning rate to adapt to the fine-tuned objective.

The main contributions are summarized as follows: (1) We propose a new UAV visual geo-localization method PVDA, which performs view distribution alignment as well as location classification in an adversarial learning framework to close the domain gap between UAV-view images and satellite-view images. (2) We develop a novel progressive adversarial learning strategy, in which the feature encoder is continuously promoted in competition with the view discriminator and able to generate location-dependent yet view-invariant features for training images as well as unseen images of new locations.

\begin{figure*}[t]
	\centering
	\includegraphics[width=1\linewidth]{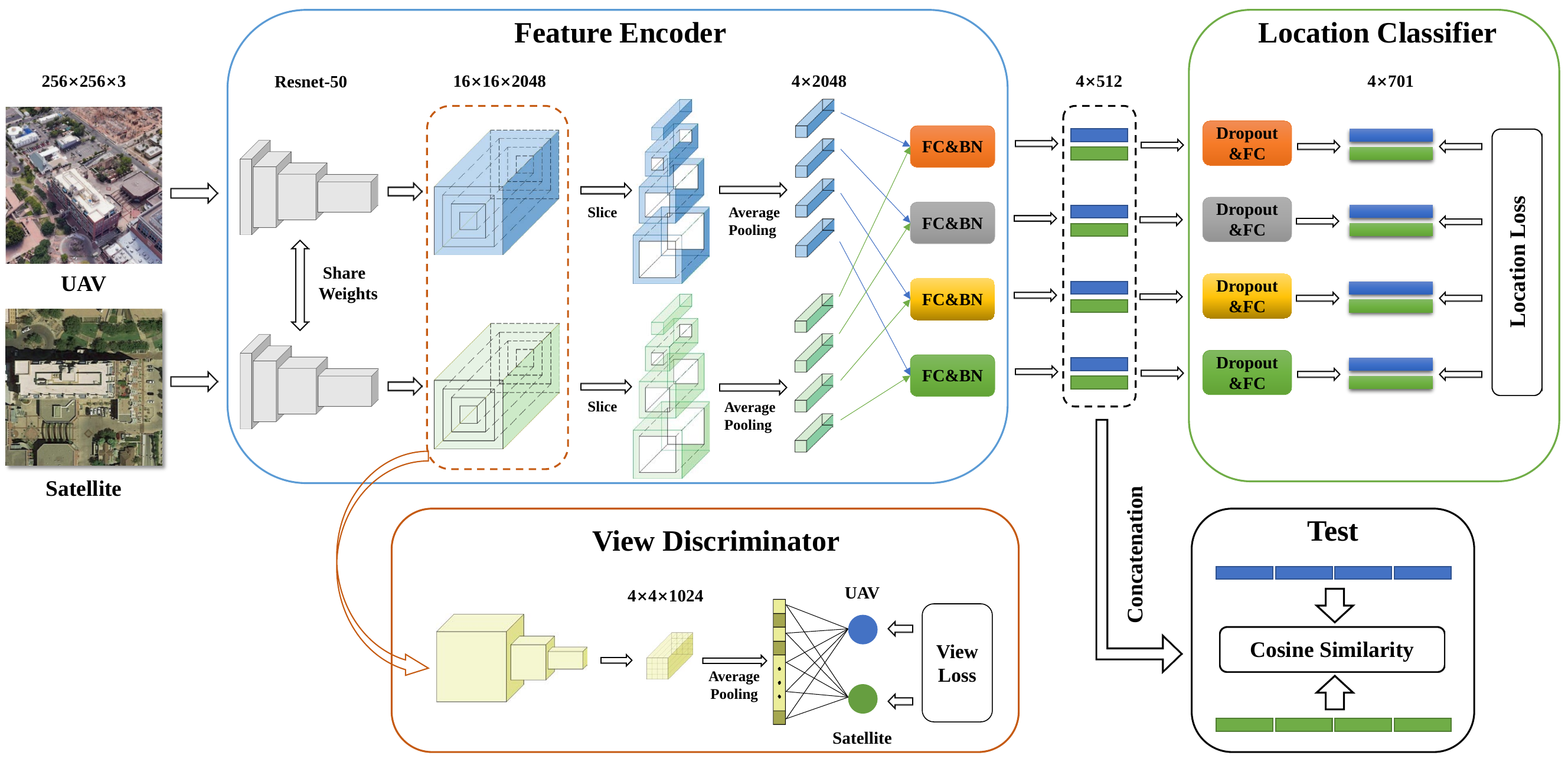}
	\vskip -5pt
	\caption{Overall framework of our method. In this exemplar, a feature encoder, a 701-way location classifier, and a 2-way view discriminator are trained on images of 701 buildings. The feature encoder takes images resized to $256 \times 256 \times 3$ as input and outputs four 512-dimensional vectors for each image.}
	\vskip -15pt
	\label{Fig:framework}
\end{figure*}

\section{Related works}

The main stream in the research for UAV-satellite geo-localization is based on the location classification framework.
Zheng et al.~\cite{zheng2020university} first formulated the UAV-satellite geo-localization and presented the University-1652 benchmark including multi-view images of different buildings.
They viewed each location as one class and employed a two-branch CNN to learn a classification model.
Ding et al.~\cite{ding2020practical} considered the imbalance of UAV-view and satellite-view images and presented a Location Classification Matching (LCM) method.
They tried to learn a location-dependent feature space and implemented the cross-view matching of images from unseen locations via feature similarity ranking.
Wang et al.~\cite{wang2021each} devised a Local Pattern Network (LPN) where a set of location classifiers are trained on part-level features generated by a square-ring feature partition strategy.
Compared to early studies~\cite{zheng2020university,ding2020practical} using global features of the whole image, LPN explores spatial contextual information around the geographic target and achieves more accurate part matching.
Lin et al.~\cite{lin2022joint} jointly performed feature learning and key-point detection to pay more attention on salient regions.
The above methods~\cite{zheng2020university,ding2020practical,wang2021each,lin2022joint} map UAV-view and satellite-view images into a common feature space and search for the classification boundaries among locations, while neglecting the gap between the UAV view and satellite view.

Tian et al.~\cite{tian2021uav} synthesized several vertical view images for a UAV-view image and then employed LPN to match the new synthetic UAV-view images against satellite-view images for geo-localization.
Although their method can decrease the viewpoint variations between the UAV view and satellite view, extra perspective projection transformation is introduced in the cross-view image matching.
Zhuang et al.~\cite{zhuang2021faster} added a multi-scale block attention mechanism into LPN for reinforcing salient features in local regions and applied the KL loss to enhance the similarity between paired UAV-view and satellite-view images.
Dai et al.~\cite{dai2022transformer} presented a transformer-based model which achieves automatic region segmentation to obtain part-level features and utilizes the triplet loss for feature alignment.
Different from the above two methods~\cite{zhuang2021faster,dai2022transformer} that focus on closing the distance between paired UAV-view and satellite-view images, our method performs global distribution alignment of the UAV view and satellite view to reduce the domain gap between them.
With the proposed progressive adversarial learning strategy, we can learn a location-dependent yet view-invariant feature space, which is crucial to matching cross-view images of unseen locations.

\section{Method}

As shown in Fig.~\ref{Fig:framework}, the proposed PVDA incorporates a feature encoder, a location classifier, and a view discriminator that learns to determine whether an image was captured by a UAV or a satellite.
The three modules are jointly optimized with a novel progressive adversarial learning strategy, where the location classifier guides the feature encoder to produce location-dependent features while competition between the feature encoder and the view discriminator enables distribution alignment of the two views.
During training, the three modules are constructed on a training dataset $D$ containing $M$ UAV-view images $\{(x_i^\mathrm{u},y_i^\mathrm{u})\}_{i=1:M}$ and $N$ satellite-view images $\{(x_j^\mathrm{s},y_j^\mathrm{s})\}_{j=1:N}$ of $C$ geographic locations, where $y_i^\mathrm{u}, y_j^\mathrm{s} \in \{1, 2, ..., C\}$ denote the location labels of the $i$-th UAV-view image and the $j$-th satellite-view image, respectively.

\subsection{Architecture of the proposed PVDA}\label{sec:framework}

\subsubsection{Feature encoder.}
ResNet-50~\cite{he2016deep} is adopted as the backbone network to extract CNN features from input images.
Following the previous work~\cite{wang2021each}, we learn a common backbone network for UAV-view and satellite-view images since they have similar patterns.
The feature maps produced by ResNet-50 are divided into multiple parts with a square-ring partition strategy~\cite{wang2021each} to separately aggregate information of the central geographic target and contextual information of surroundings.
This strategy explicitly enhances the consistency of local features of images in the same location and performs well against rotation variation as demonstrated in the previous works~\cite{wang2021each,zhuang2021faster,tian2021uav}.
Specifically, the original feature maps are partitioned into four parts, each of which is aggregated to a feature vector with an average-pooling layer.
Next, each feature vector is sent to fully connected layers and batch normalization layers for further refinement. 
Note that four branches are developed corresponding to four parts segmented by the square-ring partition strategy, so that the feature encoder can better learn the characteristics of different regions.
Finally, part-level features $\{g_{i,l}^\mathrm{u}\}_{l=1:4}$ and $\{g_{j,l}^\mathrm{s}\}_{l=1:4}$ are generated for the UAV-view image $x_i^\mathrm{u}$ and the satellite-view image $x_j^\mathrm{s}$, respectively.

\subsubsection{Location classifier.}
A four-branch classifier~\cite{wang2021each} is constructed to separate images captured at different locations based on part-level embeddings.
Concretely, embedding vectors $\{g_{i,l}^\mathrm{u}\}_{i=1:M}$ and $\{g_{j,l}^\mathrm{s}\}_{j=1:N}$ are sent to the $l$-th branch including a dropout layer and a fully connected layer with softmax operation to predict their probability distributions $\{p_{i,l}^\mathrm{u}\}_{i=1:M}$ and $\{p_{j,l}^\mathrm{s}\}_{j=1:N}$.
To evaluate the loss of location classification, we accumulate the cross-entropy between the location labels and the predicted probability distribution of each part.
This procedure can be formulated by
\begin{equation}
	\ell ^\mathrm{L} = - \sum_{i=1:M}\sum_{l=1:4} {\log {p_{i,l}^\mathrm{u}(y_i^\mathrm{u})}} - \sum_{j=1:N}\sum_{l=1:4} {\log {p_{j,l}^\mathrm{s}(y_j^\mathrm{s})}},
	\label{eq:locationloss}
\end{equation}
where $p_{i,l}^\mathrm{u}(y_i^\mathrm{u})$ denotes the probability of the ground-truth location of image $x_i^\mathrm{u}$ and $p_{j,l}^\mathrm{s}(y_j^\mathrm{s})$ indicates the probability of the ground-truth location of image $x_j^\mathrm{s}$.

\begin{figure}[t]
	\centering
	\includegraphics[width=0.67\linewidth]{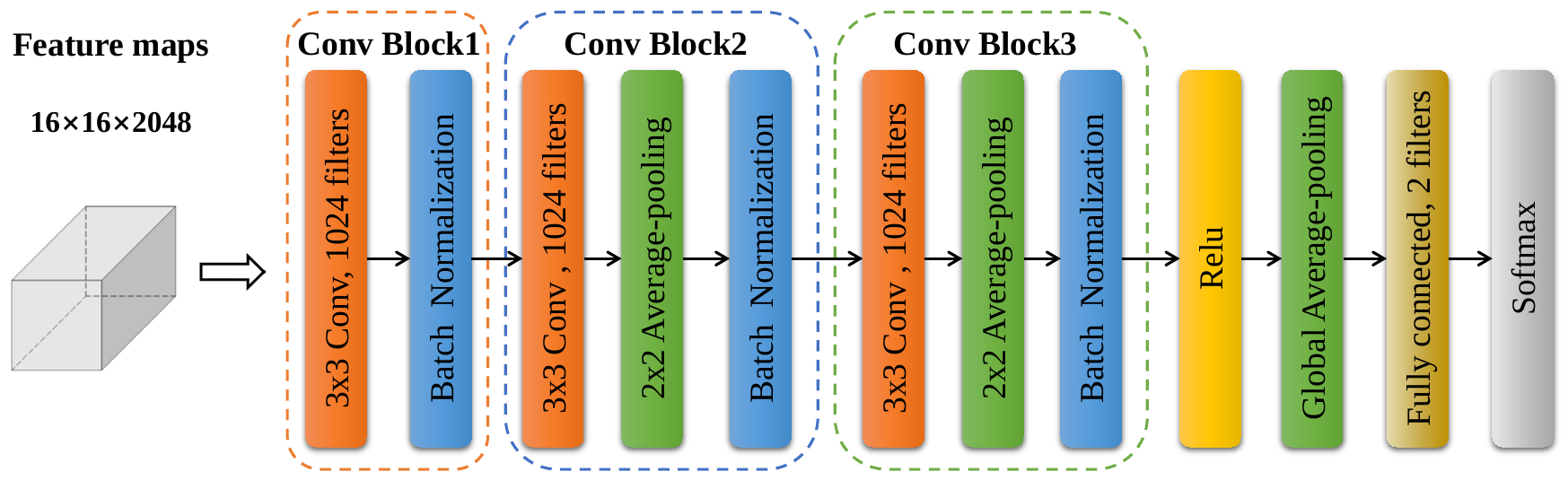}
	\vskip -10pt
	\caption{Architecture of the view discriminator.
	}
	\vskip -15pt
	\label{Fig:discriminator}
\end{figure}

\subsubsection{View discriminator.}

We construct the view discriminator on the intermediate feature maps output by ResNet-50 rather than the part-level embedding vectors.
The main reason is that spatial details in the intermediate feature maps are critical to differentiate between UAV-view images and satellite-view images.
Moreover, the intermediate features are enforced to be view-invariant by the progressive adversarial learning strategy, then the subsequent part-level embeddings derived from them can also get such characteristics.

As illustrated in Fig.~\ref{Fig:discriminator}, the view discriminator first refines the feature maps with three convolutional blocks (denoted as Conv Block 1 to Conv Block 3) followed by the ReLU nonlinearity.
Convolutional layers in Conv Block 2 and Conv Block 3 are equipped with spatial pooling to reduce the resolution of produced feature maps.
The view discriminator ends with a global average pooling layer for aggregating spatial information and a two-way fully connected layer with softmax for view prediction.
Given feature maps $f_i^\mathrm{u}$ of a UAV-view image $x_i^\mathrm{u}$, the view discriminator outputs a 2D vector $q_i^\mathrm{u} = [q_{i,1}^\mathrm{u}, q_{i,2}^\mathrm{u}]$ indicating the probabilities that the input belongs to the UAV view and the satellite view, respectively.
Similarly, a 2D probability vector $q_j^\mathrm{s}=[q_{j,1}^\mathrm{s},q_{j,2}^\mathrm{s}]$ is generated for a satellite-view image $x_j^\mathrm{s}$.
The loss to train the view discriminator is the cross-entropy between the output probabilities and the ground-truth view labels formulated as
\begin{equation}
	\ell ^\mathrm{V} = - \sum_{i=1:M} {\log {q_{i,1}^\mathrm{u}}} - \sum_{j=1:N} {\log {q_{j,2}^\mathrm{s}}},
	\label{eq:viewloss}
\end{equation}
where $q_{i,1}^\mathrm{u}$ is the predicted probability of the UAV view for a UAV-view image and $q_{j,2}^\mathrm{s}$ denotes the probability of the satellite view for a satellite-view image.

\subsection{Progressive adversarial learning strategy} \label{sec:learning}

The location classifier aims to separate images of different locations and the view discriminator determines whether an input is from the UAV view or the satellite view.
It is clear that the feature encoder agrees with the location classifier on generating location-dependent features, yet it must fool the view discriminator so as to produce view-invariant features.
Therefore, we introduce an adversarial loss $\ell ^\mathrm{A}$ between the feature encoder and the view discriminator.
\begin{equation}
	\ell ^\mathrm{A} = - \sum_{i=1:M} {\log {q_{i,2}^\mathrm{u}}} - \sum_{j=1:N} {\log {q_{j,1}^\mathrm{s}}},
	\label{eq:adloss}
\end{equation}
where $q_{i,2}^\mathrm{u}$ and $q_{j,1}^\mathrm{s}$ are probabilities predicted by the view discriminator.
To be specific, $q_{i,2}^\mathrm{u}$ stands for the probability of the satellite view for a UAV-view image, and $q_{j,1}^\mathrm{s}$ denotes the probability of the UAV view for a satellite-view image.
Obviously, the adversarial loss $\ell ^\mathrm{A}$ calculates the cross-entropy between the output probability distributions and view labels opposite to the ground-truth labels so that the generated features can deceive the view discriminator into regarding UAV-view images as satellite-view images and vice versa.

During training, the three modules are iteratively optimized in the following two steps.
First, the feature encoder and the location classifier are fixed and the view discriminator is optimized with the objective function defined in Eq.~\ref{eq:viewloss}.
Second, we freeze the parameters of the view discriminator and update the feature encoder together with the location classifier by an optimization objective $\ell ^\mathrm{FL}$, which is a weighted combination of the location classification loss $\ell^L$ and an adversarial loss $\ell ^A$. More specifically,
\begin{equation}
	\ell ^\mathrm{FL} = \ell^\mathrm{L} + \alpha \cdot \ell ^\mathrm{A},
	\label{eq:loss}
\end{equation}
where $\alpha$ is a weight balancing the two optimization objectives.

In the early stage of training, the three modules cannot fit the training data well, so minimizing the location classification loss $\ell^\mathrm{L}$ is just as important as the adversarial loss $\ell ^\mathrm{A}$.
With the continuous update of parameters, they are getting stronger, so it is easier to achieve location classification but harder for the feature encoder to deceive the view discriminator.
Inspired by these observations, the proposed progressive adversarial learning strategy gradually puts the emphasis on optimizing the adversarial loss $\ell ^\mathrm{A}$ by increasing the weight $\alpha$ in Eq.~\ref{eq:loss} in regular intervals.
Motivated by the work~\cite{2016SGDR}, we integrate the warm restart technique that periodically restarts and decays the learning rate into the proposed progressive adversarial learning strategy.
Especially, the learning rate is restarted whenever the weight of the adversarial loss is increased.
This learning mechanism enables fast gradient descent with a big learning rate when the optimization objective in Eq.~\ref{eq:loss} is fine-tuned by the increasing weight and employs a small learning rate to approach an optimum.

\subsection{Cross-view image matching}\label{sec:matching}

Once the feature encoder is learned, it is evaluated on cross-view image retrieval tasks with query/gallery data captured at new locations.
The classifier cannot predict unseen locations of query/gallery images since they have their own location label space independent to the training images.
Nonetheless, we can match a query image against candidate images in the gallery by comparing their embeddings in the learned feature space.
Specifically, a query image is sent to the feature encoder to acquire part-level embedding vectors, which are then concatenated to form the query representation.
Likewise, we can get the representation of a candidate image by concatenating its part-level embedding vectors produced by the feature encoder.
The final retrieval results are derived from the ranking of feature similarity measured by the cosine similarity.

\section{Experiments}
\subsection{Dataset and experimental settings}

We employ the large-scale University-1652 dataset to evaluate our method on two tasks, i.e., UAV-to-satellite image matching for UAV-view target localization and satellite-to-UAV image matching for UAV navigation.
This dataset consists of images of 1,652 buildings from 72 universities and provides one geo-tagged satellite-view image and 54 UAV-view images for each building.
The UAV-view images were captured in a simulation flight, and there exist large scale variations as well as rotation variations in this dataset.
All the buildings are divided into two parts: 701 buildings from 33 universities for training and 951 buildings from the rest 39 universities for testing.
For UAV-to-satellite image matching, there are 37,855 UAV-view queries and 951 satellite-view candidates including 701 true-matched images and 250 distractors.
For satellite-to-UAV image matching, we have 701 satellite-view queries and 51,355 UAV-view candidates including 37,855 true-matched images and 13,500 distractors.
Following the previous work~\cite{zheng2020university}, we employ Recall@K and average precision (AP) as evaluation metrics.

ResNet-50 is employed as the backbone of the feature encoder and initialized with the pre-trained weights on ImageNet~\cite{deng2009imagenet}.
We take the feature maps produced by the fifth convolution block as image representations and remove the down-sampling operation in the fifth convolution block to retain more details.
The feature encoder and the location classifier are learned with an SGD optimizer and the view discriminator is learned with an Adam optimizer.
The parameter $\alpha$ in Eq.~\ref{eq:loss} is initialized to 0.9 and increased by 0.1 every 140 epochs.
We set the initial learning rate to 0.001, 0.01, 0.01, and 0.002 for ResNet-50, the rest layers in the feature encoder, the location classifier, and the view discriminator, respectively.
The learning rate is decayed by multiplying 0.8 after 60 epochs and 120 epochs, and reset to the initial value whenever $\alpha$ is increased.

\begin{table}[t]
	\caption{Cross-view image matching accuracy of different methods.}
	\vskip -10pt
	\centering
	\label{tab:compare_SOAT}
	\begin{tabular}{|c|cc|cc|cc|cc|}
		\hline
		\multirow{2}{*}{Method}  & \multicolumn{4}{c|}{256 $\times$ 256} & \multicolumn{4}{c|}{384 $\times$ 384}\\
		\cline{2-9} &  \multicolumn{2}{c|}{UAV-to-satellite} & \multicolumn{2}{c|}{Satellite-to-UAV}
		&\multicolumn{2}{c|}{UAV-to-satellite} & \multicolumn{2}{c|}{Satellite-to-UAV}\\
		\cline{2-9} & Recall@1 & AP & Recall@1 & AP  & Recall@1 & AP & Recall@1 & AP  \\
		\hline
		Zheng et al.~\cite{zheng2020university}  &58.49\% &63.31\% &71.18\% &58.74\% &62.99\% &67.69\% &75.75\% &62.09\% \\
		LCM~\cite{ding2020practical} & -- & -- & -- &-- & 66.65\% &70.82\% &79.89\% &65.38\% \\
		LPN~\cite{wang2021each}&75.93\%  &79.14\%  &86.45\% &74.79\% &78.02\% &80.99\% &86.16\% &76.56\% \\
		USAM~\cite{lin2022joint} &77.60\% &80.55\% &	86.59\% &	75.96\%  & -- & -- & -- &-- \\
		PCL~\cite{tian2021uav} & 79.47\% &83.63\% &87.69\% &78.51\% &81.63\% &85.46\% &89.73\% &80.84\%  \\
		MSBA~\cite{zhuang2021faster}& 82.33\% &84.78\% &90.58\% &81.61\% &86.61\% &88.55\% &92.15\% &84.45\% \\
		FSRA~\cite{dai2022transformer} &82.25\% &84.82\% &87.87\% &81.53\% &84.82\% &87.03\% &87.59\% &83.37\% \\
		\hline
		PVDA   &   82.73\%  & 85.19\%  &  92.30\%  &  82.48\%  & 87.34\%  & 89.26\%  &  93.72\%  &  86.04\% \\
		\hline
	\end{tabular}
	\vskip -15pt
\end{table}

\subsection{Experimental results}
\subsubsection{Comparison to the state-of-the-arts.}
The proposed PVDA is compared with the existing UAV visual geo-localization methods and the results of using two input image sizes (i.e., 256 and 384) are shown in Table~\ref{tab:compare_SOAT}.
ResNet-50 is employed as the backbone in CNN-based methods~\cite{zheng2020university,ding2020practical,wang2021each,zhuang2021faster,tian2021uav,lin2022joint}, while FSRA~\cite{dai2022transformer} takes the Vision Transformer~\cite{2021vit} as the backbone. 
All the comparison methods pre-train their backbones on ImageNet~\cite{deng2009imagenet}.

As shown in Table~\ref{tab:compare_SOAT}, the proposed PVDA performs better than the other methods on both UAV-to-satellite image matching and satellite-to-UAV image matching using two image sizes.
Specifically, our method significantly outperforms methods~\cite{zheng2020university,ding2020practical} using global image features.
LPN~\cite{wang2021each}, PCL~\cite{tian2021uav}, and our method adopt the same feature encoder and location classifier.
The difference is that PCL~\cite{tian2021uav} adopts UAV-to-satellite image synthesis as data augmentation while our method carries out distribution alignment of UAV-view images and satellite-view images in an adversarial learning framework.
In comparison to LPN~\cite{wang2021each} and PCL~\cite{tian2021uav}, our method has achieved superior performance, which demonstrates the effectiveness of our learning framework.
USAM~\cite{lin2022joint} and MSBA~\cite{zhuang2021faster} are also based on LPN~\cite{wang2021each}.
The former forces the feature encoder to focus on salient regions by embedding two attention modules in ResNet-50, while the latter incorporates both global features and local features.
FSRA~\cite{dai2022transformer} adopts a transformer-based backbone, which is stronger than ResNet-50.
Nonetheless, our method still achieves better performance than USAM~\cite{lin2022joint}, MSBA~\cite{zhuang2021faster}, and FSRA~\cite{dai2022transformer}.
It is worth noting that the inference time of our method is the same as that of LPN~\cite{wang2021each}.
However, PCL~\cite{tian2021uav}, USAM~\cite{lin2022joint}, MSBA~\cite{zhuang2021faster}, and FSRA~\cite{dai2022transformer} need more memory usage and longer inference time due to image synthesis~\cite{tian2021uav} or the more complex network structure~\cite{lin2022joint,zhuang2021faster,dai2022transformer}.
Therefore, our method is very competitive in real-world applications considering the limitation on computing resources and time.

\begin{table}[t]
	\caption{UAV-view target localization accuracy with multiple queries.}
	\vskip -10pt
	\centering
	\label{tab:multiquery}
	\begin{tabular}{|c|ccc|ccc|}
		\hline
		\multirow{2}{*}{Method}  & \multicolumn{3}{c|}{256 $\times$ 256} & \multicolumn{3}{c|}{384 $\times$ 384}\\
		\cline{2-7} &  \multicolumn{3}{c|}{UAV-to-satellite} &\multicolumn{3}{c|}{UAV-to-satellite} \\
		\cline{2-7} & Recall@1 & Recall@5 & AP  & Recall@1 & Recall@5  & AP  \\
		\hline
		Zheng et al.~\cite{zheng2020university}  & 69.33\% &86.73\%  &73.14\% &-- &--  &--\\
		LCM~\cite{ding2020practical} & -- & -- & --  & 77.89\% & 91.30\%  & 81.05\%\\
		\hline
		PVDA (Single-query)    &82.73\% & 93.60\%  &85.19\% &87.34\% & 95.78\% &  89.26\% \\
		PVDA (Multi-query)    &92.01\% & 98.00\% &93.30\%  &95.29\% & 98.43\% &  95.99\% \\
		\hline
	\end{tabular}
	\vskip -15pt
\end{table}

\subsubsection{Multi-query image matching.}

In the above experiments, the models take a single UAV-view image as query and return a ranking list of candidate satellite-view images in the UAV-view target localization task.
Due to the fact that the UAV may capture multiple images of one geographic target from different viewpoints and heights in the flight, it is natural to localize the target under a multi-query setting.
To this end, we conduct experiments under two image sizes to investigate the effectiveness of using multiple UAV-view images as queries.
Particularly, we retrieve the most relevant satellite-view image in the gallery according to the average of features of 54 UAV-view queries.
Experimental results shown in Table~\ref{tab:multiquery} verify that more accurate target localization can be achieved by integrating multiple queries into a more comprehensive description.
The improvements on AP are 8.11\% and 6.73\% for image sizes $256 \times 256$ and $384 \times 384$, respectively.
Compared with other geo-localization methods~\cite{zheng2020university,ding2020practical}, our method also performs best under the multi-query setting, which further demonstrates the superiority of our method.

\subsubsection{Visualization of UAV visual geo-localization results.}

\begin{figure*}[t]
	\centering
	\includegraphics[width=0.85\linewidth]{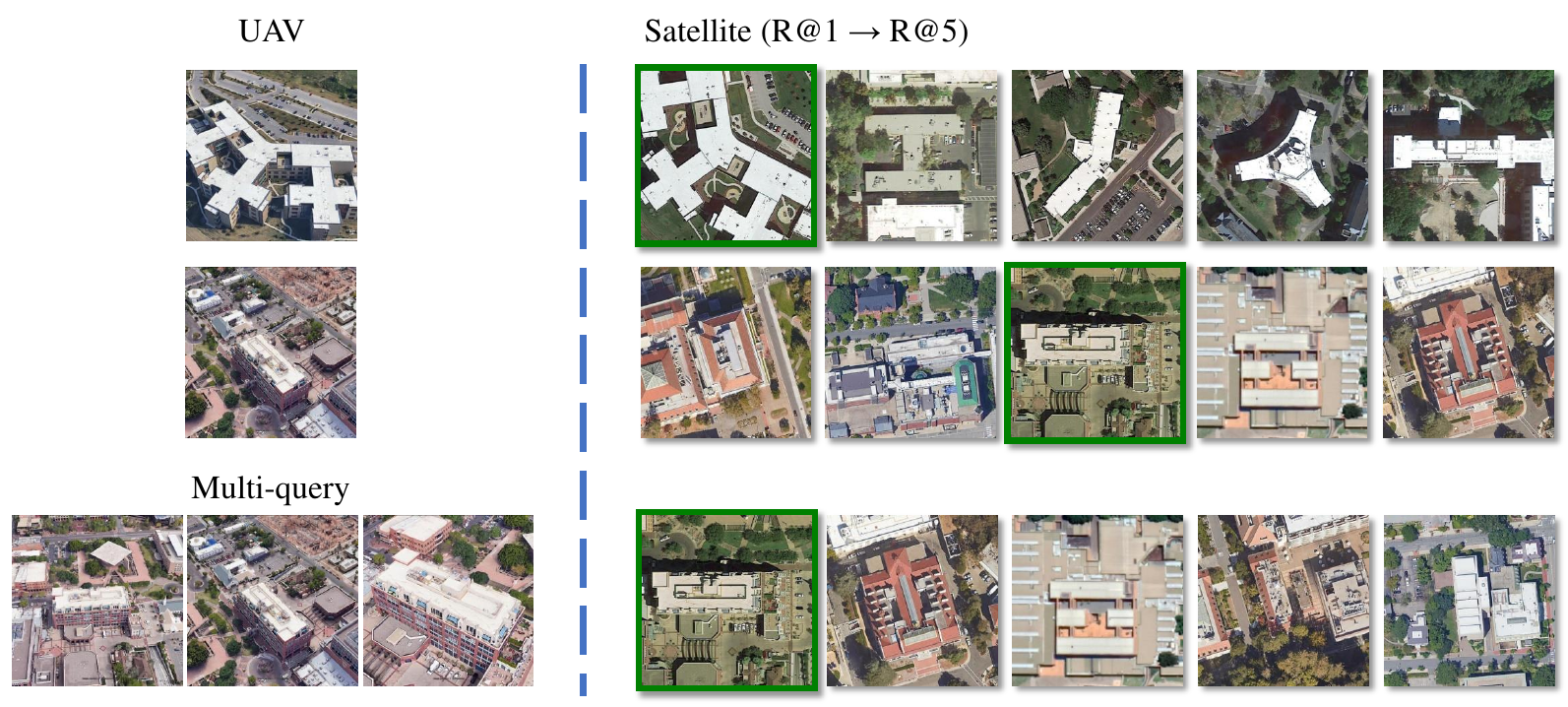}
	\vskip -10pt
	\caption{Top-5 retrieved satellite-view images in the UAV-view target localization task. The first and second rows display the matching results with a single query, while the third row employs multiple UAV-view images as queries. The true-matched satellite-view images are annotated with green borders.
	}
	\vskip -5pt
	\label{Fig:visualization1}
\end{figure*}

\begin{figure*}[t]
	\centering
	\includegraphics[width=0.7\linewidth]{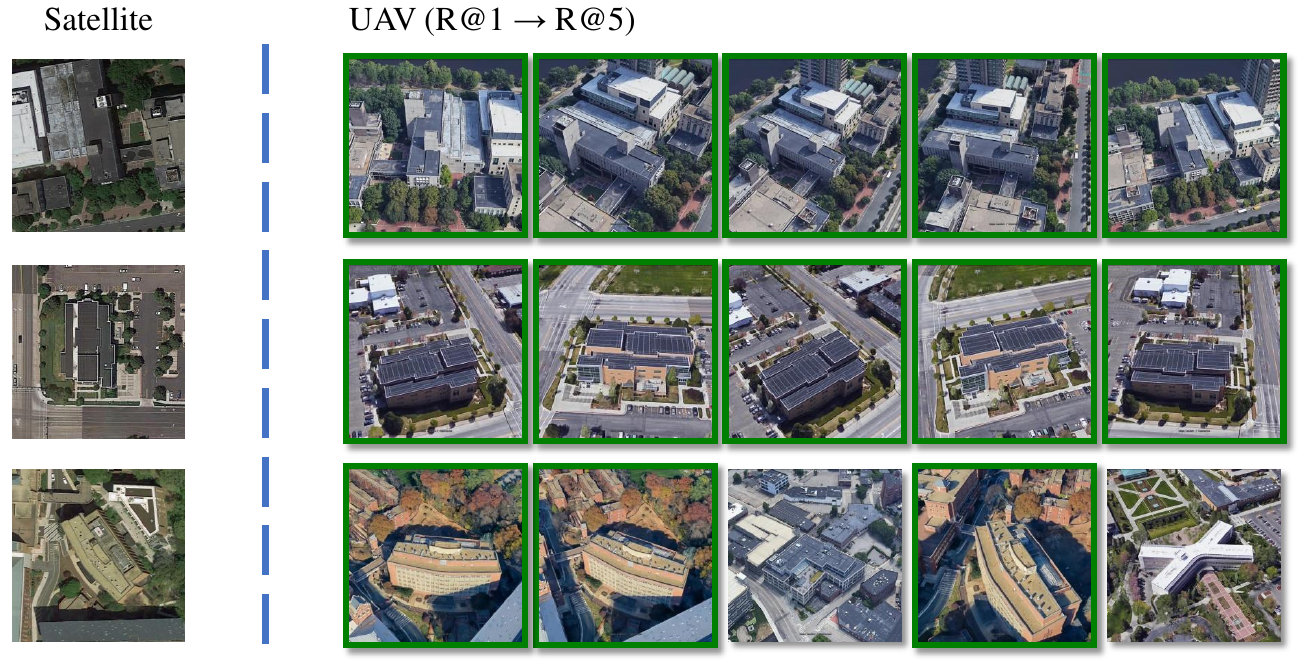}
	\vskip -10pt
	\caption{Top-5 retrieved UAV-view images in the UAV navigation task. The true-matched UAV-view images are annotated with green borders.
	}
	\vskip -10pt
	\label{Fig:visualization2}
\end{figure*}

Some cross-view image retrieval results of the proposed method on the UAV-view target localization task and the UAV navigation task are visualized in Fig.~\ref{Fig:visualization1} and Fig.~\ref{Fig:visualization2}, respectively.
Input images are resized to $256 \times 256$.
In the UAV-view target localization task, there is only one true-matched satellite-view image in the gallery for each UAV-view query.
As shown in Fig.~\ref{Fig:visualization1}, our method can correctly retrieve the satellite-view image for the first query though there exist distinct appearance variations between the UAV-satellite image pair.
The second row displays a failure case where some irrelevant surroundings of the target building (e.g., the long road on the right top of the query image) interfere with the image matching.
When multiple UAV-view images are used as queries, our method can extract more accurate representations of this target and recall the true-matched satellite-view image in top-1 as shown in the third row.
In the UAV navigation task, there are multiple true-matched UAV-view images in the gallery for each satellite-view query.
As shown in Fig.~\ref{Fig:visualization2}, our method is able to recall UAV-view images of the target building even in large viewpoint and scale variations.

\subsection{Ablation studies}
To further study whether the proposed progressive adversarial learning strategy is beneficial to parameter optimization, we design two baseline strategies for comparison.
The first baseline adopts a constant weight $\alpha$ (see Eq.~\ref{eq:loss}) and periodically restarts the learning rate, so the adversarial loss will not be progressively emphasized over time. 
The second baseline utilizes an increasing weight $\alpha$ like the proposed learning strategy, but doesn't simulate warm restart of the learning rate when weight $\alpha$ is increased.
To be specific, the learning rate is initialized in accordance with our learning strategy and decayed by multiplying 0.8 after executing 140, 280, and 420 epochs.
As shown in Table~\ref{tab:Ablation}, the proposed progressive adversarial learning strategy outperforms the first baseline by progressively emphasizing the adversarial loss over time.
In comparison to the second baseline, our method also achieves superior performance with periodic warm restart of the learning rate.
These observations reveal that the proposed progressive adversarial learning strategy is effective in parameter optimization and the warm restart scheme works well with the evolving objective of adversarial learning.

\begin{table}[t] 
	\begin{center}
		\caption{Comparison of image matching accuracy between our method and baselines.}
		\vskip -5pt
		\label{tab:Ablation}
		\begin{tabular}{|c|cc|cc|}
			\hline
			\multirow{2}{*}{Method} &  \multicolumn{2}{c|}{UAV-to-satellite} & \multicolumn{2}{c|}{Satellite-to-UAV}\\
			\cline{2-5} & Recall@1 & AP & Recall@1 & AP  \\
			\hline
			constant $\alpha$, w/ warm restart  & 82.00\% & 84.55\% &  91.73\% &81.96\% \\
			increasing $\alpha$, w/o warm restart & 81.54\% & 84.09\% &  92.30\% &81.05\% \\
			\hline
		increasing $\alpha$, w/ warm restart (Ours) & 82.73\%  & 85.19\%  &  92.30\%  &  82.48\%\\
			\hline
		\end{tabular}
	\end{center}
	\vskip -15pt
\end{table}

\section{Conclusion}
In this paper, we have presented an end-to-end learning framework to alleviate the distribution shift between UAV-view images and satellite-view images in UAV visual geo-localization.
A novel progressive adversarial learning strategy is developed to perform view distribution alignment and location classification in a common feature space.
By doing this, our method can learn location-dependent yet view-invariant features and thus achieves better performance than the existing methods on the large-scale University-1652 dataset.

\bibliographystyle{splncs04_unsort}
\bibliography{geo}

\end{document}